\documentclass{article}

\PassOptionsToPackage{numbers, compress}{natbib}

\usepackage[preprint, nonatbib]{nips_2018}

\usepackage[utf8]{inputenc} 
\usepackage[T1]{fontenc}    
\usepackage{hyperref}       
\usepackage{url}            
\usepackage{booktabs}       
\usepackage{amsfonts}       
\usepackage{nicefrac}       
\usepackage{microtype}      

\usepackage{graphicx}
\usepackage{subfigure}
\usepackage{booktabs} 

\usepackage{amsmath}
\usepackage{amssymb}
\usepackage{color}
\usepackage{natbib}
\usepackage{makecell}
\usepackage{cuted}
\usepackage{comment}

\usepackage{tikz}
\usetikzlibrary{arrows,shapes,backgrounds,through,shadows}

\tikzstyle{cont}=[circle, draw,thick,minimum size=7.5mm,line width=1pt,>=stealth]  

\tikzstyle{obs}=[fill=blue!10,thick]  

\tikzstyle{contobs}+=[cont]
\tikzstyle{contobs}+=[obs]
\tikzstyle{discobs}+=[disc]
\tikzstyle{discobs}+=[obs]

\tikzstyle{dgraph}=[->, line width=1.5pt]

\DeclareGraphicsExtensions{.pdf}

\def\be{\begin{equation}}
\def\ee{\end{equation}}

\newcommand{\av}[1]{\langle{#1}\rangle}
\newcommand{\KL}[2]{D_\text{KL}\left[#1|| #2\right]}

\title{Improving latent variable descriptiveness with AutoGen}

\author{
  Alex Mansbridge \\
  University College London \\
  Alan Turing Institute \\
  \And
  Roberto Fierimonte \\
  University College London \\
  \AND
  Ilya Feige \\
  ASI Data Science \\
  \And
  David Barber \\
  University College London \\
  Alan Turing Institute \\
  }

\begin{document}

\maketitle

\begin{abstract}
Powerful generative models, particularly in Natural Language Modelling, are commonly trained by maximizing a variational lower bound on the data log likelihood.  These models often suffer from poor use of their latent variable, with ad-hoc annealing factors used to encourage retention of information in the latent variable. We discuss an alternative and general approach to latent variable modelling, based on an objective that combines the data log likelihood as well as the likelihood of a perfect reconstruction through an autoencoder. Tying these together ensures by design that the latent variable captures information about the observations, whilst retaining the ability to generate well. Interestingly, though this approach is a priori unrelated to VAEs, the lower bound attained is identical to the standard VAE bound but with the addition of a simple pre-factor; thus, providing a formal interpretation of the commonly used, ad-hoc pre-factors in training VAEs.
\end{abstract}

\section{Introduction}
\label{sec:intro}

Generative latent variable models are probabilistic models of observed data $x$ of the form $p(x,z)=p(x|z)p(z)$, where $z$ is the latent variable. These models are widespread in machine learning and statistics. They are useful both because of their ability to generate new data and because the posterior $p(z|x)$ provides insight into the low dimensional representation $z$ corresponding to the high dimensional observation $x$.  These latent $z$ values are then often used in downstream tasks, such as topic modelling \cite{dieng2016topicrnn}, multi-modal language modeling \cite{kiros2014multimodal}, and image captioning \cite{mansimov2015generating, pu2016variational}. 

Latent variable models, particularly in the form of  Variational Autoencoders (VAEs) \cite{kingma2014stochastic, rezende2014stochastic}, have been successfully employed in natural language modelling tasks using varied architectures for both the encoder and the decoder \cite{bowman2016generating, dieng2016topicrnn, semeniuta2017hybrid, yang2017improved, shah2017generating}. However, an architecture that is able to effectively capture meaningful semantic information into its latent variables is yet to be discovered. 

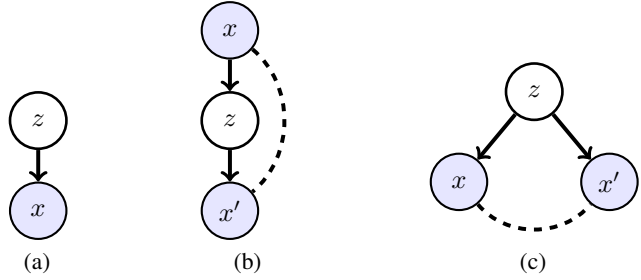
\begin{figure}[t]
\begin{center}
\subfigure[]{\begin{tikzpicture}[dgraph]
\node[cont] (z) at (0,1.2) {$z$};
\node[contobs] (x) at (0,0) {$x$};
\draw(z)--(x);
\end{tikzpicture}
\label{fig:graphical_standard_generative}}
\hspace{1.5cm}
\subfigure[]{\begin{tikzpicture}[dgraph]
\node[contobs] (x) at (0,2.4) {$x$};
\node[cont] (z) at (0,1.2) {$z$};
\node[contobs] (xp) at (0,0) {$x'$};
\draw(z)--(xp);\draw(x)--(z);
\draw[-,dashed](x) to [bend left=50] (xp);
\end{tikzpicture}
\label{fig:graphical_autoencoder}}
\hspace{1.5cm}
\subfigure[]{\begin{tikzpicture}[dgraph]
\node[cont] (z) at (0,1.2) {$z$};
\node[contobs] (x) at (-1,0) {$x$};
\node[contobs] (xp) at (1,0) {$x'$};
\draw(z)--(xp);\draw(z)--(x);
\draw[-,dashed](x) to [bend right=50] (xp);
\end{tikzpicture}
\label{fig:graphical_autogen}}
\end{center}
\caption{(a) Standard generative model. (b) Stochastic autoencoder with tied observations. (c) Equivalent tied stochastic autoencoder with AutoGen parameterisation.}
\end{figure}

A ``Standard VAE'' approach to language modelling was given by \cite{bowman2016generating}, the graphical model for which is shown in Figure~\ref{fig:graphical_standard_generative}.  This forms a generative model $p_\theta(x|z)p(z)$ of sentence $x$, based on latent variable $z$, where $\theta$ are the parameters of the generative model. Since the integral $p(x) = \int p_\theta(x|z)p(z)dz$ is typically intractable, a common approach is to maximize the Evidence Lower Bound (ELBO) on the log likelihood,
\be
\label{eq:ELBO}
\log p(x) \geq \av{\log p_\theta(x|z)} - \KL{q(z|x)}{p(z)}
\ee
where expectation $\av{\cdot}$ is with respect to the variational ``encoder'' $q(z|x)$, and $\KL{\cdot}{\cdot}$ represents the Kullback-Leibler (KL) divergence. Summing over all datapoints $x$ gives a lower bound on the likelihood of the full dataset.  

In language modelling, typically both the generative model (``decoder'') $p(x|z)$, and variational distribution (``encoder'') $q(z|x)$, are parameterised using an LSTM recurrent neural network -- see for example \cite{bowman2016generating}. This generative model -- combined with the highly structured teacher-forcing training technique -- is so powerful that the maximum ELBO is achieved without making appreciable use of the latent variable in the model. Indeed, if trained using the SGVB algorithm \cite{kingma2014stochastic}, the model learns to ignore the latent representation and effectively relies solely on the decoder to generate good sentences. This is evidenced by the KL term in the objective function converging to zero, indicating that the approximate posterior distribution of the latent variable is trivially converging to its prior distribution. 

The dependency between what is represented by latent variables, and the capacity of the decoding distribution (i.e., its ability to model the data without using the latent) is a general phenomenon. \cite{yang2017improved} used a lower capacity dilated CNN decoder to generate sentences, preventing the KL term going to zero. \cite{gulrajani2016pixelvae, higgins2016beta} have discussed this in the context of image processing. A clear explanation of this phenomenon in terms of Bit-Back Coding is given in \cite{chen2016variational}.

A mechanism to avoid the model ignoring the latent entirely, while allowing a high capacity decoder is discussed in \cite{bowman2016generating} and uses an alternative training procedure called ``KL annealing''  -- slowly turning on the KL term in the ELBO during training. KL annealing allows the model to use its latent variable to some degree by forcing the model into a local maximum of its objective function. Modifying the training procedure in this way to preferentially obtain local maxima suggests that the objective function used in \cite{bowman2016generating} may not be ideal for modelling language in such a way as to create a model that leverages its latent variables.

\section{Training generative models with AutoGen}
\label{sec:autogen}

We propose a new generative latent-variable model inspired by the autoencoder framework. Autoencoders are trained to reconstruct data through a low-dimensional bottleneck layer, and as a result, construct a dimensionally-reduced representation from which the data can be reconstructed. By encouraging reconstruction in our model, we force the latent variable to represent the input data, overcoming the issues in \cite{bowman2016generating} of the latent variable being ignored, as discussed in Section~\ref{sec:intro}.

However, using an autoencoder alone does not enable generation from a prior distribution, as in the case of VAEs. To leverage both generation as well as high-fidelity reconstruction from the latent variable, we propose to maximize the likelihoods of both:
\be
\label{eq:autogen_obj}
\mathcal{L}_\text{AutoGen} \;=\; \sum_{n} \;\;
\underbrace{\log p(x=x_n)}_{\text{generation (VAE)}} \,+\, 
\underbrace{\log p(x'=x_n | x=x_n)}_{\text{reconstruction (autoencoder)}}
\ee
where $x'$ represents the reconstruction and the training data is denoted by $\{x_n\}$. Thus the input data $x$ and the output $x'$ are tied, much like an autoencoder. Crucially, optimizing $\mathcal{L}_\text{AutoGen}$ does not correspond to optimizing the log likelihood of the data, nor would a lower bound on $\mathcal{L}_\text{AutoGen}$ correspond to the ELBO used in VAEs, due to the addition of the autoencoder term. Instead, $\mathcal{L}_\text{AutoGen}$ represents the log likelihood of different model that combines both VAEs and autoencoders.

To see this, we develop AutoGen further by writing the autoencoding term as a stochastic autoencoder:
\be
\label{eq:stochastic_AE}
p(x'=x_n | x=x_n) = \int p(x'=x_n |z)\,p(z| x=x_n)\, dz
\ee
which encourages high-fidelity reconstruction from its stochastic embedding $z$. The graphical model associated with this reconstruction term alone is shown in Figure~\ref{fig:graphical_autoencoder}. Similarly, the generation term in Eq.~\eqref{eq:autogen_obj}, can be chosen to be a generative model as in the case of VAEs:
\be
\label{eq:standard_generative}
p(x=x_n) = \int p(x=x_n |z)\,p(z)\, dz
\ee
The graphical model associated with this generative term is shown in Figure~\ref{fig:graphical_standard_generative}. 

As yet, we have not specified how Eqs.~\eqref{eq:stochastic_AE} and \eqref{eq:standard_generative}, the two terms in $\mathcal{L}_\text{AutoGen}$, connect to each other. To do so, we make two assumptions: firstly, we assume that the generative model $p(x=x_n |z)$ is the same as the reconstruction model $p(x'=x_n |z)$ in the stochastic autoencoer. The second assumption is that the encoding and decoding distributions in the stochastic autoencoder are symmetric. Using Bayes' rule, we write this symmetry assumption as
\be
\label{eq:autogen_sym}
p(z|x=x_n) = \frac{p(x'=x_n|z)p(z)}{p(x=x_n)}
\ee

These two assumptions constrain the two otherwise-independent models, allowing Autogen to demand both generation from the prior as in a VAE and high-fidelity reconstructions from the latent variable as in an autoencoder, all while specifying essentially one single probability model, $p(x=x_n|z)$. 

The graphical representation of AutoGen is shown in Figure~\ref{fig:graphical_autogen}, where the dashed line corresponds to the tying (equality) of the input and output of the autoencoder. Indeed, with these assumptions, $\mathcal{L}_\text{AutoGen}$ can be written as:
\be
\label{eq:autogen_joint}
\mathcal{L}_\text{AutoGen} = \sum_{n} \;
\log p(x=x_n) + 
\log p(x'=x_n | x=x_n)
= \sum_{n} \;\log p(x'=x_n, x=x_n)
\ee
which is why the graphical model can be interpreted as the tying of two separate generations from the same model $p(x=x_n|z)$.

With the AutoGen assumptions, a simple lower bound for Eq.~\eqref{eq:autogen_obj} can be derived following standard arguments:
\begin{align}
\label{eq:autogen_elbo}
\mathcal{L}_\text{AutoGen} \;\geq\; \sum_{n} \;\;
2 \, \av{\log p(x'=x_n | z)}_{q(z|x_n)} 
\,-\, \KL{q(z|x_n)}{p(z)} 
\end{align}
where we write the approximate posterior as $q(z|x'=x_n,x=x_n) = q(z|x_n)$ for brevity. A detailed derivation of Eq.~\eqref{eq:autogen_elbo} is presented in Section~\ref{sec:derivation}; the reader can skip this derivation without losing the flow of the presentation.

\subsection{Derivation of the lower bound}
\label{sec:derivation}

To derive the AutoGen lower bound in Eq.~\eqref{eq:autogen_elbo}, we begin by constructing a variational lower bound on the stochastic autoencoder term in $\mathcal{L}_\text{AutoGen}$, see Eqs.~\eqref{eq:autogen_obj} and \eqref{eq:stochastic_AE}. In what follows we suppress the sum over the data $\{x_n\}$ for clarity. Specifically, we write,
\begin{align}
\mathcal{L}_\text{AutoGen} &= \log p(x=x_n) + \log \int dz \, p(x'=x_n | z) p(z | x=x_n)\\
&= \log p(x=x_n) + \log \int dz \, q(z|x_n) \frac{p(x'=x_n | z) p(z | x=x_n)}{q(z|x_n)}
\notag
\end{align}
where $q(z|x'=x_n,x=x_n) = q(z|x_n)$ is the variational approximate posterior. Using Jensen's inequality, we get a lower bound on the objective function:
\begin{align}
\mathcal{L}_\text{AutoGen} &\geq \log p(x=x_n) + \int dz \, q(z|x_n) \log \frac{p(x'=x_n | z) p(z | x=x_n)}{q(z|x_n)} \\
&= \int dz \, q(z|x_n) \log \frac{p(x'=x_n | z) p(z | x=x_n) p(x=x_n)}{q(z|x_n)} 
\notag
\end{align}
The symmetry hypothesis of AutoGen in Eq.~\eqref{eq:autogen_sym} then gives,
\begin{align}
\mathcal{L}_\text{AutoGen} &\geq \int dz \; q(z|x_n) \log \frac{p(x'=x_n | z)^2 p(z)}{q(z|x_n)} \\
&= 2 \, \av{\log p(x'=x_n | z)}_{q(z|x_n)} - \KL{q(z|x_n)}{p(z)}
\notag
\end{align}
Hence we have shown Eq.~\eqref{eq:autogen_elbo}.


\subsection{Discussion of AutoGen}

We see that the variational lower bound derived for AutoGen in Eq.~\eqref{eq:autogen_elbo} is the same as that of the Standard VAE \cite{kingma2014stochastic, rezende2014stochastic}, but with a factor of 2 in the reconstruction term. It is important to emphasize, however, that the Autogen objective is not a lower bound on the data log likelihood. Maximizing the lower bound in Eq.~\eqref{eq:autogen_elbo} represents a criterion for training a generative model $p(x|z)$ that evenly balances both good spontaneous generation of the data $p(x=x_n)$ as well as high-fidelity reconstruction $p(x'=x_n|x=x_n)$, as it is a lower bound on the sum of those log likelihoods, Eq.~\eqref{eq:autogen_obj}. 

Of course, AutoGen does not force the latent variable to encode information in a particular way (e.g., semantic representation in language models), but it is a necessary condition that the latent represents the data well in order to reconstruct it. We discuss the relation between AutoGen and other efforts to influence the latent representation of VAEs in Section~\ref{sec:discussion}.


A natural generalisation of the AutoGen objective and assumptions, see Eq.~\eqref{eq:autogen_joint}, would be to maximize the joint with $m$ independent-but-tied reconstructions, instead of just 2. Following the arguments in Section~\ref{sec:derivation} leads to a lower bound with a factor of $1+m$ in front of the generative term: 
\begin{align}
\label{eq:autogen_elbo_m}
\mathcal{L}_\text{AutoGen} (m) 
&= \log  p(x^1=x_n, \ldots, x^m=x_n, x=x_n) \\
&\geq (1+m) \,
\big\langle \log p(x_n | z)\big\rangle_{q(z|x_n)}
- \KL{q(z|x_n)}{p(z)} 
\notag
\end{align}
Larger $m$ encourages better reconstructions at the expense of poorer generation. We discuss the impact of the choice of $m$ in Section~\ref{sec:exp}.

\section{Experiments}
\label{sec:exp}

We train four separate language models, all based on the implementation of \cite{bowman2016generating}. We train two variants of this model using the regular ELBO - one such variant uses KL annealing, and the other does not. We refer to these variants as ``Standard VAEs''. We train our baseline AutoGen model using the objective in Eq.~\eqref{eq:autogen_elbo}, and train an AutoGen variant using the objective in Eq.~\eqref{eq:autogen_elbo_m} with $m=2$.

All of the models were trained using the BookCorpus dataset \cite{zhu2015aligning}, which contains sentences from a collection of 11,038 books. We restrict our data to contain only sentences with length between 5 and 30 words, and restrict our vocabulary to the most common 20,000 words. We use 90\% of the data for training and 10\% for testing. After preprocessing, this equates to 58.8 million training sentences and 6.5 million test sentences. All models in this section are trained using word drop as in \cite{bowman2016generating}.

Neither AutoGen models are trained using KL annealing. We consider KL annealing as an unprincipled approach, as it destroys the relevant lower bound during training. In contrast, AutoGen provides a unfettered lower bound throughout training, albeit a lower bound on $\log p(x'=x_n,x=x_n)$, rather than the data log likelihood $\log p(x=x_n)$. Despite this, we consider AutoGen only to be useful if it improves the descriptiveness of the latent variable as compared to the Standard VAE with annealing, hence we compare to the Standard VAE without and with KL annealing.

\subsection{Optimization results}
\label{sec:exp_learning_curves}

\begin{figure*}[t]
\begin{center}
    \includegraphics[width=.49\linewidth]{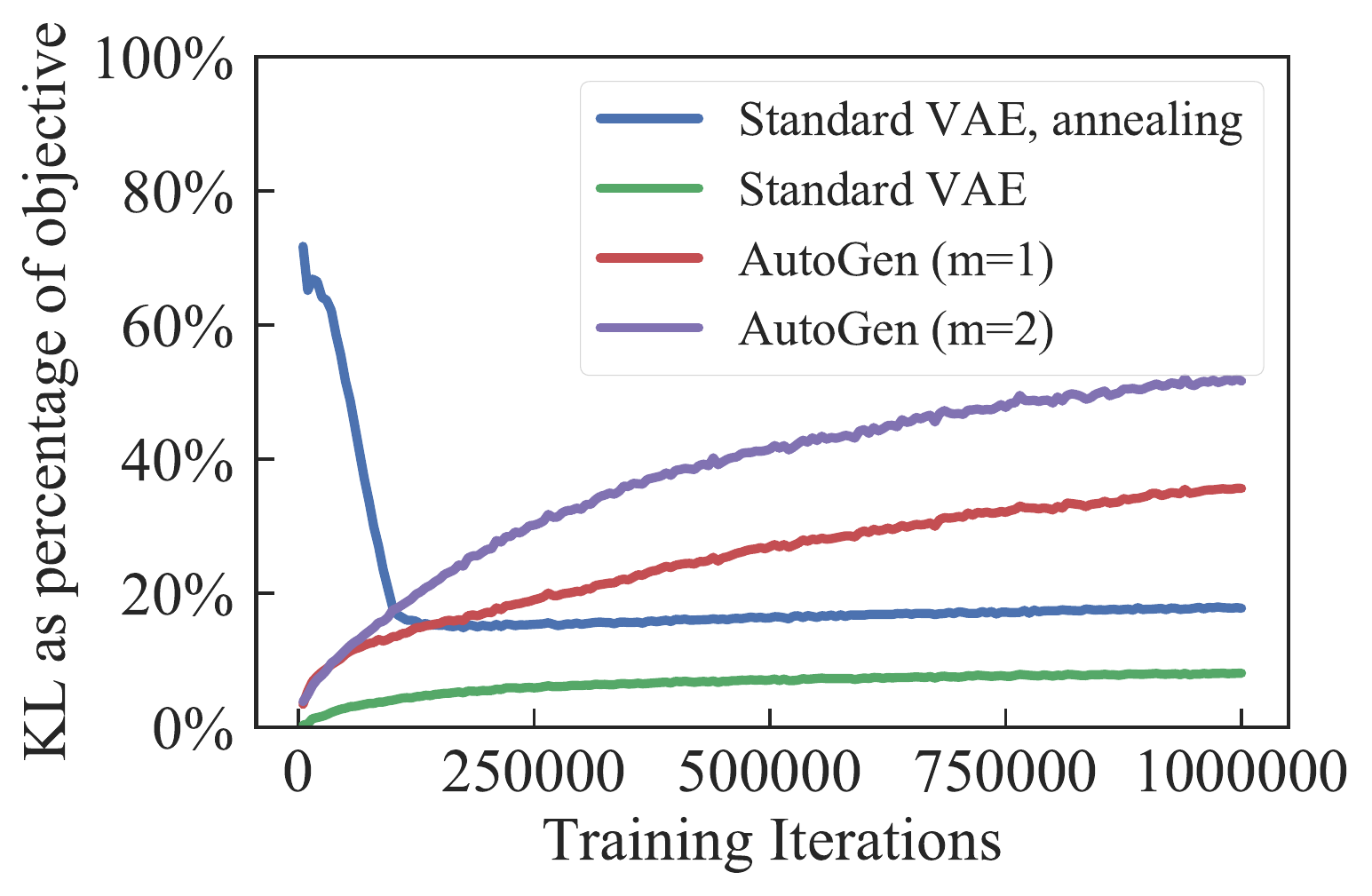}
    \includegraphics[width=.49\linewidth]{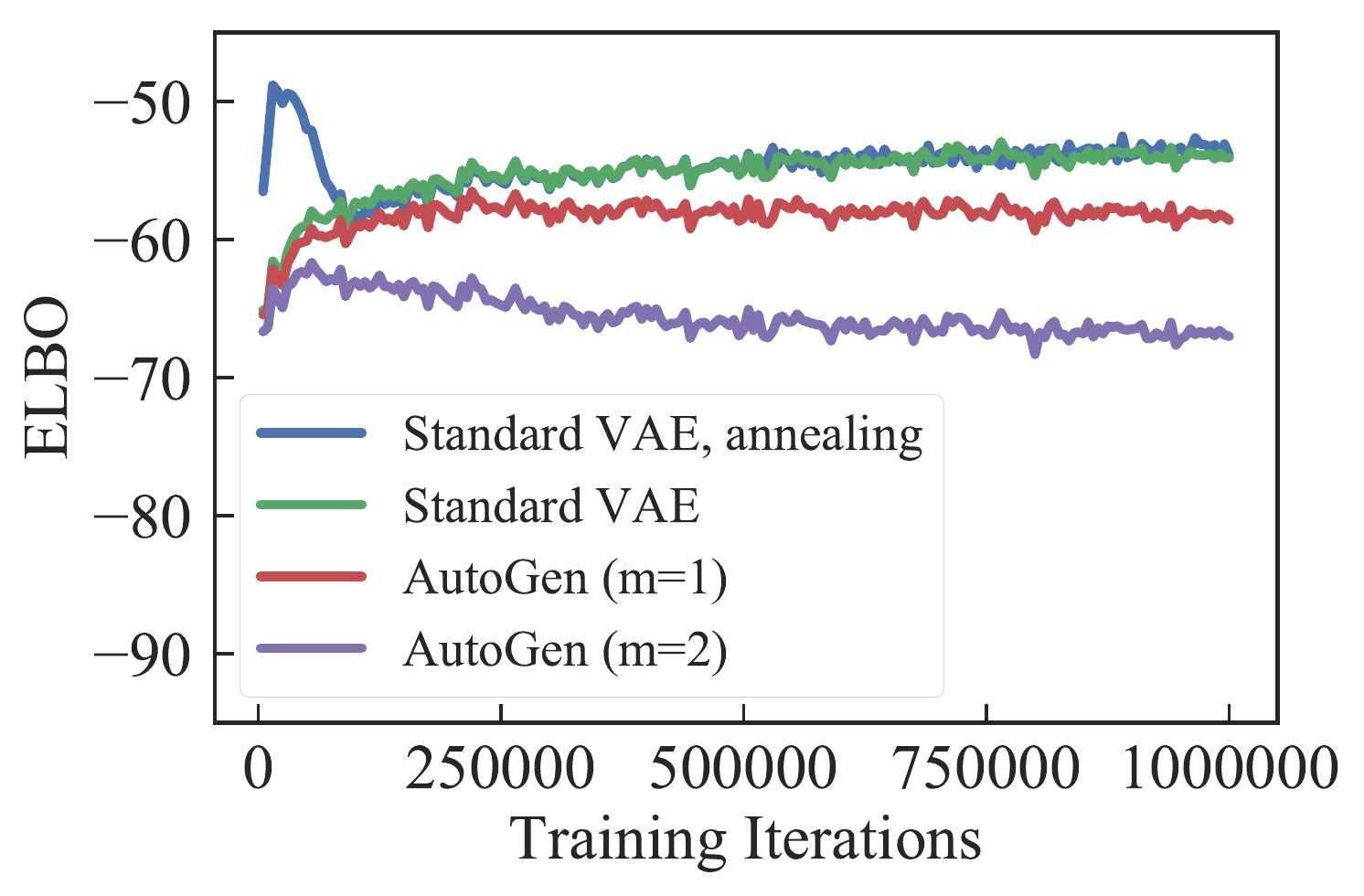}
\end{center}
    \caption{(Left) $-\KL{q(z|x_n)}{p(z)}$ term as a \% of overall objective for the four models throughout training. (Right) ELBO (log likelihood lower bound, Eq.~\eqref{eq:ELBO}) for the four models throughout training.}
    \label{fig:optimisation_plots}
\end{figure*}

We train all models for 1 million iterations using mini-batches of 200 sentences. The objective functions differ between the four models, and so it is not meaningful to directly compare them. Instead, in Figure~\ref{fig:optimisation_plots} (left), we show the \% of the objective function that is accounted for by the KL term. Despite the fact that AutoGen has a larger pre-factor in front of the $\langle \log p(x|z) \rangle_{q(z|x)}$ term, the KL term becomes more and more significant with respect to the overall objective function for AutoGen with $m=1$ and $m=2$, as compared to the Standard VAE. 
This suggests that the latent in AutoGen is putting less emphasis on matching the prior distribution, and more emphasis on directly representing the data.

To understand the impact of AutoGen on the log likelihood of the training data (which is only one of two terms in the AutoGen objective, Eq.~\eqref{eq:autogen_obj}), we can compare the Standard VAE ELBO in Eq.~\eqref{eq:ELBO} of the four models during training. Since the ELBO is the objective function for the Standard VAE, we expect it to be a relatively tight lower bound on the log likelihood. However, this only applies to the Standard VAE. Indeed, if the ELBO for AutoGen is similar to that of the Standard VAE, we can conclude that the AutoGen model is approximately concurrently maximizing the log likelihood as well as its reconstruction-specific objective function. 

In Figure~\ref{fig:optimisation_plots} (right) we show the ELBO for all four models. We see that, though the baseline AutoGen ($m=1$) ELBO is below that of the Standard VAE, it tracks the Standard VAE ELBO well and is non-decreasing. On the other hand, for the more aggressive AutoGen with $m=2$, the ELBO starts decreasing early on in training and continues to do so as its objective function is maximized. Thus, for the baseline AutoGen with objective function corresponding to maximizing Eq.~\eqref{eq:autogen_obj}, we expect decent reconstructions without significantly compromising generation from the prior, whereas AutoGen ($m=2$) may have a much more degraded ability to generate well. In Sections \ref{sec:sentencerecon} and \ref{sec:sentencegen} we corroborate this expectation qualitatively by studying samples from the models.

\subsection{Sentence reconstruction}
\label{sec:sentencerecon}

Indications that AutoGen should more powerfully encode information into its latent variable were given theoretically in the construction of AutoGen in Section~\ref{sec:autogen} as well as in Section~\ref{sec:exp_learning_curves} from the optimization results. To see what this means for explicit samples, we perform a study of the sentences reconstructed by the Standard VAE as compared to those by the AutoGen model.

In Table~\ref{tab:posterior}, an input sentence $x$ is taken from our test set, and a reconstruction $x'$ is presented that maximizes $p(x'|z)$, as determined using beam search. We sample $z\sim q(z|x)$ in this process, meaning we find different reconstructions every time from the same input sentence, despite the beam search procedure in the reconstruction.

\begin{table*}[t]
\caption{Reconstructed sentences from the Standard VAE and AutoGen. Sentences are not ``cherry picked'': these are the first four sentences reconstructed from a grammatically correct input sentence, between 4 and 20 words in length (for aesthetics), and with none of the sentences containing an unknown token (for readability).}
\label{tab:posterior}
\begin{center}
\fontsize{9}{9.5}
\begin{sc}
\begin{tabular}{ p{.23\textwidth} p{.21\textwidth} p{.22\textwidth} p{.22\textwidth} }
\toprule
Input Sentence 
& VAE Reconstruction
& VAE Reconstruction (annealing)
& AutoGen Reconstruction ($m=1$) 
\\ \midrule
``more or less?''
&``oh yeah.'' ``
&``what about you?''
&``more or less?''
\\ \hline
why would you need to talk when they can do it for you?
&how could n't i ?&why do you want to know if i can find out of here?
&why would you need to know if you can do it for you?
\\ \hline
she had no idea how beautiful she truly was .& she hadn't .
& she had no idea what she was talking about .
&she had no idea how beautiful she was to .
\\ \hline
``i guess some people never learn.''
&``i love you.
& `` you know what you 're thinking.''
&``i guess our parents never exist.
\\
\bottomrule
\end{tabular}
\end{sc}
\end{center}
\vskip -0.1in
\end{table*}

AutoGen is qualitatively better at reconstructing sentences than the Standard VAE. Indeed, even when the input sentence is not reconstructed verbatim, AutoGen is able to generate a coherent sentence with a similar meaning by using semantically similar words. For example in the last sentence, by replacing ``some people'' with ``our parents'', and ``never learn'' with ``never exist''. On the other hand, the Standard VAE reconstructions regularly produce sentences that have little relation to the input. Note that without annealing, the Standard VAE regularly ignores the latent, producing short, high-probability sentences reconstructed from the prior.

To make these results more quantitative, we ran three versions of a survey in which respondents were asked to judge the best reconstructions from two models. In the first survey, we received responses from 6 people who compared 120 pairs of reconstructions from the Standard VAE and the Standard VAE with annealing. The second survey received responses from 13 people over 260 sentences and compared reconstructions from the Standard VAE with annealing to AutoGen ($m=1$). The third compared AutoGen ($m=1$) to AutoGen ($m=2$) and received 23 responses over 575 sentences. None of the respondents in these surveys were authors of this paper. The surveys were designed in this way to provide an easy binary question for the respondents. They provide a suitable test of the models due to the transitive nature of the comparisons.

Our survey results are shown in Table~\ref{tab:reconstruction_survey}. We can clearly see that AutoGen with $m=2$ outperforms AutoGen with $m=1$, as expected. Similarly, AutoGen with $m=1$ outperforms the Standard VAE with annealing, and the Standard VAE with annealing outperforms the Standard VAE . All results have greater than 99\% confidence.

\begin{table}[h]
\caption{Results from a blind survey comparing reconstruction quality. Respondents were asked to ``choose the best reconstruction'', and where ambiguous, could discard reconstruction pairs.}
\label{tab:reconstruction_survey}
\begin{center}
\begin{small}
\begin{sc}
\begin{tabular}{ lc }
\toprule
\makecell[l]{Model~1 vs. model~2}
& 
\makecell[l]{\% of responses with model~1 as winner} \\
\midrule
\makecell[l]{VAE (annealing) vs. VAE}
& 66\% \\
\makecell[l]{AutoGen ($m=1$) vs. VAE (annealing) }
& 88\% \\
\makecell[l]{AutoGen ($m=2$) vs. AutoGen ($m=1$) }
\hspace{0.5cm}
& 88\% \\ 
\bottomrule
\end{tabular}
\end{sc}
\end{small}
\end{center}
\vskip -0.1in
\end{table}

\subsection{Sentence generation}
\label{sec:sentencegen}

The objective function of AutoGen encourages the generation of higher-fidelity reconstructions from its approximate posterior. The fundamental trade-off is that it may be less capable of generating sentences from its prior. 

To investigate the qualitative impact of this trade-off, we now generate samples from the prior $z \sim \mathcal{N}(0,\text{I})$ of the Standard VAE and AutoGen. For a given latent $z$, we generate sentences $x'$ as in Section~\ref{sec:sentencerecon}. Results are shown in Table~\ref{tab:prior}, where we see that both models appear to generate similarly coherent sentences; there appears to be no obvious qualitative difference between the Standard VAE and AutoGen. 

\begin{table*}
\caption{Sentences generated from the prior, $z \sim \mathcal{N}(0,\text{I})$, for the Standard VAE and AutoGen. Sentences are not ``cherry picked'': they are produced in the same way as those in Table~\ref{tab:posterior}.}
\label{tab:prior}
\begin{center}
\fontsize{9}{9.5}
\begin{sc}
\begin{tabular}{p{.23\textwidth} p{.35\textwidth} p{.33\textwidth}}
\toprule
VAE Generation 
& VAE Generation (annealing)  
& AutoGen Generation ($m=1$) \\
\midrule
the only thing that mattered.
&she just looked up.
&they don't show themselves in mind , or something to hide.
\\ \hline
he gave her go.
&she felt her lips together.
&her eyes widen, frowning.
\\ \hline
``good morning,'' i thought.
&my hands began to fill the void of what was happening to me.
&the lights lit up around me.
\\ \hline
she turned to herself.
&at first i knew he would have to.
&i just feel like fun.
\\
\bottomrule
\end{tabular}
\end{sc}
\end{center}
\vskip -0.1in
\end{table*}

To be more quantitative, we ran a survey of 23 people -- none of which were the authors -- considering 392 sentences generated from the priors of all four of the models under consideration. We applied the same sentence filters to these generated sentences as we did to those generated in Table~\ref{tab:prior}. We then asked the respondents whether or not a given sentence ``made sense'', maintaining the binary nature of the question, but allowing the respondent to interpret the meaning of a sentence ``making sense''. To minimize systematic effects, each respondent saw a maximum of 20 questions, evenly distributed between the four models. All sentences in the surveys were randomly shuffled with the model information obfuscated.


The results of our survey are shown in Table~\ref{tab:generation_survey}. Since the Standard VAE generates systematically shorter sentences than the training data, which are inherently more likely to be meaningful, we split our results into short and long sentences (with length $\leq10$ and $>10$ tokens, respectively). We conclude that the Standard VAE with annealing is better at generating short sentences than AutoGen ($m=1$). However, both models achieve equal results on generation quality for longer sentences. We also see that AutoGen ($m=2$) generates significantly worse sentences than other models, as expected. All results that differ by more 1 percentage point in the table are statistically significant with confidence greater than $99\%$. 

\begin{table}
\caption{Results from a blind survey testing generation quality. Respondents were asked ``does this sentence make sense'' for a randomized list of sentences evenly sampled from the four models. Results are split into two sentence lengths $L$ in order to mitigate the bias of the Standard VAE models to generate short sentences.}
\label{tab:generation_survey}
\begin{center}
\begin{small}
\begin{sc}
\begin{tabular}{lcc}
\toprule
\makecell[l]{Model}
& \makecell[l]{\% meaningful ($L\leq10$)}
& \makecell[l]{\% meaningful ($L>10$)}
\hspace{0.5cm}
\\
\midrule
VAE
&75\%
&N/A
\\ 
VAE (annealing)
&76\%
&32\%
\\ 
AutoGen ($m=1$)
&50\%
&32\%
\\ 
AutoGen ($m=2$)
&29\%
&5\%
\\ \bottomrule
\end{tabular}
\end{sc}
\end{small}
\end{center}
\vskip -0.1in
\end{table}

\subsection{Latent manifold structure}
\label{sec:latent-manifold-results}

Finally, with high-fidelity reconstructions from the latent, one would expect to be able to witness the smoothness of the latent space well. This seems to be the case, as can be seen in Table~\ref{tab:autogen-latent-posterior-interpolation}, where we show the reconstructions of a linear interpolation between two encoded sentences for Standard VAE with annealing and for AutoGen ($m=1$). The AutoGen interpolation seems to be qualitatively smoother, in the sense that, while neighbouring sentences are more similar, there are fewer instances of reconstructing the same sentences at subsequent interpolation steps.

\begin{table*}
\caption{Latent variable interpolation. Two sentences, $x_1$ and $x_2$ (first and last sentences in the table) are randomly selected from the test dataset, which provide $z_i \sim q(z|x_i)$. Sentences are then generated along 10 evenly spaced steps from $z_1$ to $z_2$. This interpolation was not ``cherry picked'': this was our first generated interpolation; we use the same sentence filters as all previous tables.}
\label{tab:autogen-latent-posterior-interpolation}
\begin{center}
\begin{small}
\begin{sc}
\begin{tabular}{ p{5.5cm} p{5cm} }
\toprule 
VAE (annealing) &  AutoGen ($m=1$) \\
\midrule
``i'll do anything, blake.'' 
&``i'll do anything, blake.''
\\ 
``i'll be right back then.''
& ``i'll do it, though.''
\\ 
``i'll tell me like that.''
& ``i'll say it, sir.''
\\ 
i dont know what to say.
& ``i've done it once.''
\\ 
i dont know what to say.
& i dont think that was it.
\\ 
i dont think about that way.
& i wish so, though.
\\ 
i'll be right now.
& i bet it's okay.
\\ 
i was so much.
& i know how dad.
\\ 
i looked at him.
& i laughed at jack.
\\ 
i looked at him.
& i looked at sam.
\\ 
i looked at adam.
& i looked at adam. 
\\
\bottomrule
\end{tabular}
\end{sc}
\end{small}
\end{center}
\vskip -0.1in
\end{table*}

The reconstructions from the Standard VAE without annealing have little dependence on the latent, and AutoGen ($m=2$) struggles to generate from the prior. As a consequence, both of these models show highly non-smooth interpolations with little similarity between subsequent sentences. The results for these models have therefore been omitted.

We have provided only a single sample interpolation, and though it was not cherry picked, we do not attempt to make a statistically significant statement on the smoothness of the latent space. Given the theoretical construction of AutoGen, and the robust results shown in previous sections, we consider smoothness to be expected. The sample shown is consistent with our expectations, though we do not consider it a definite empirical result.

\section{Discussion}
\label{sec:discussion}

We have seen that AutoGen successfully improves the fidelity of reconstructions from the latent variable as compared to VAEs. It does so in a principled way, by adding the likelihood of a perfect reconstruction to the objective function of the standard VAE, namely the log likelihood of the data. 

This is especially useful in VAE models where the decoding distribution is very powerful, such as the autoregressive RNN used in \cite{bowman2016generating}. We note that we continue to use (word) dropout, as in \cite{bowman2016generating}, with AutoGen because it improves both the baseline VAE models, as well as the AutoGen models. We postulate that dropout would not be needed if teacher forcing was not used in our experiments, but leave that study to future work as we believe that our experiments are sufficient to show the impact of AutoGen in a controlled way.

Other work toward enabling latent variables in VAE models to learn meaningful representations has focused on managing the structure of the representation, such as ensuring disentanglement. A detailed discussion of disentanglement in the context of VAEs is given in \cite{higgins2016beta} and its references. An example of disentangling representations in the context of image generation is \cite{gulrajani2016pixelvae}, where the authors restrict the decoding model to describe only local information in the image (e.g., texture, shading), allowing their latent variables to describe global information (e.g., object geometry, overall color).

Demanding high-fidelity reconstructions from latent variables in a model (e.g., AutoGen) is in tension with demanding specific information to be stored in the latent variables (e.g., disentanglement). This can be seen very clearly by comparing our work to \cite{higgins2016beta}, where the authors introduce a factor of $\beta$ in front of the KL-divergence term of the Standard VAE objective function, the ELBO. They find that $\beta > 1$ is required to improve the disentanglement of their latent representations.

Interestingly, $\beta > 1$ corresponds analytically to $-1<m<0$ in Eq.~\eqref{eq:autogen_elbo_m}, since the overall normalization of the objective function does not impact the location of its extrema. That is,
\be
(1+m) \; \big\langle \log p(x | z)\big\rangle_{q(z|x)}
- \KL{q(z|x)}{p(z)} \;\Longleftrightarrow\;
\big\langle \log p(x | z)\big\rangle_{q(z|x)}
- \beta\, \KL{q(z|x)}{p(z)}
\notag
\ee
with $\beta = (1+m)^{-1}$. 

Since $m$ in AutoGen represents the number of times a high-fidelity reconstruction is demanded in the objective function (in addition to a single generation from the prior), $\beta$-VAE with $\beta > 1$ is analytically equivalent to demanding a \emph{negative} number of high-fidelity reconstructions. As an analytic function of $m$, with larger $m$ corresponding to higher-fidelity reconstructions, negative $m$ would correspond to a deprecation of the reconstruction quality. This is indeed what the authors in \cite{higgins2016beta} find and discuss. They view $\beta$-VAE as a technique to trade off more disentangled representations at the cost of lower-fidelity reconstructions, in contrast to our view of AutoGen as a technique to trade off higher-fidelity reconstructions at the cost of slightly inferior generation from the prior.

In connecting to $\beta$-VAE, we have considered AutoGen with $m$ as a real number. Practically, $m$ need not take on integer values, and we imagine that for some tasks it may be beneficial to tune $m>0$ as a hyperparameter. From our results, we expect $m\approx1$ to be a useful ballpark value, with smaller $m$ improving generation from the prior, and larger $m$ improving reconstruction fidelity. The advantage of tuning $m$ as described is that it has a highly principled interpretation at integer values; namely that of demanding $m$ exact reconstructions from the latent, as derived in Section~\ref{sec:autogen}.

In this light, KL annealing amounts to starting with $m=\infty$ at the beginning, and smoothly reducing $m$ down to $0$ during training. Thus, it is equivalent to optimizing the AutoGen lower bound given in Eq.~\eqref{eq:autogen_elbo_m} with varying $m$ during training. However, AutoGen should never require KL annealing.


\section{Conclusions}
\label{sec:conc}

In this paper, we introduced AutoGen: an novel modelling approach to improving the descriptiveness of latent variables in VAEs by adding the log likelihood of $m$ high-fidelity reconstructions to the objective function. This approach is theoretically principled in that it retains a bound on a meaningful objective, and computationally amounts to a simple factor of $(1+m)$ in front of the reconstruction term in the standard ELBO. We find that the most natural version of AutoGen (with $m=1$) provides significantly better reconstructions than the Standard VAE approach to language modelling, and only minimally deprecates generation from the prior.

\section{Acknowledgments}

This work was supported by the Alan Turing Institute under the EPSRC grant EP/N510129/1 and by AWS Cloud Credits for Research.

\bibliographystyle{bib/icml2018}
\bibliography{bib/bibliography-nlp.bib}

\end{document}